\documentclass[english,12pt]{article}

\usepackage{algorithm}
\usepackage{algorithmic}
\usepackage{amsfonts}
\usepackage{amsmath}
\usepackage{amssymb}
\usepackage{amsthm}
\usepackage{bbm}
\usepackage{bm}
\usepackage{cleveref}
\usepackage{comment}
\usepackage[T1]{fontenc}
\usepackage{enumerate}
\usepackage{graphicx}
\usepackage{lmodern}
\usepackage{makeidx}
\usepackage{mathrsfs}
\usepackage{mathtools}
\usepackage{natbib}
\usepackage{placeins}
\usepackage{times}
\usepackage[normalem]{ulem}
\usepackage{xcolor}
\usepackage{xparse}

\crefname{assumption}{Assumption}{Assumptions}
\crefname{proposition}{Proposition}{Propositions}
\crefname{theorem}{Theorem}{Theorems}
\crefname{section}{Section}{Section}
\crefname{lemma}{Lemma}{Lemma}
\crefname{algorithm}{Algorithm}{Algorithms}
\crefname{example}{Example}{Examples}
\crefname{figure}{Figure}{Figure}
\crefname{appendix}{Appendix}{Appendix}
\crefname{equation}{equation}{equation}

\def\RR{{\mathbb R}}

\newcommand{\A}{\mathcal{A}}

\newcommand{\N}{\mathrm{Normal}}

\newcommand{\iid}{\stackrel{\rm iid}{\sim}} 
\newcommand{\ind}{\stackrel{\rm ind}{\sim}}

\renewcommand{\RR}{\mathbb{R}}

\def\c{\mathsf{c}}

\newtheorem{remark}{Remark}

\def\rf{\mathrm{ref}}
\def\optim{\mathrm{optim}}

\DeclareMathOperator*{\argmin}{arg\,min}

\def\bY{\mathbf{Y}}
\def\bX{\mathbf{X}}
\def\dist{\mathsf{dist}}

\def\lle{\mathrm{lle}}
\def\activ{\varphi}
\def\permutation{\sigma}
\def\indicator{\mathbb{I}}

\def\error{\varepsilon}
\def\hcauchy{\mathrm{half}\text{-}\mathrm{Cauchy}}

\begin{document}

\title{Bayesian neural networks and dimensionality reduction}

\author{
Deborshee Sen$^{1,2}$\footnotemark[1]\footnote{Corresponding author; {deborshee.sen@duke.edu}}
\and Theodore Papamarkou$^{3,4}$ 
\and David Dunson$^1$ 
}
\date{$^1$Department of Statistical Science, Duke University \\
$^2$The Statistical and Applied Mathematical Sciences Institute (SAMSI) \\
$^3$Computational Sciences and Engineering Division, Oak Ridge National Lab \\
$^4$Department of Mathematics, University of Tennessee}

\maketitle

\begin{abstract}
In conducting non-linear dimensionality reduction and feature learning, it is common to suppose that the data lie near a lower-dimensional manifold. A class of model-based approaches for such problems includes latent variables in an unknown non-linear regression function; this includes Gaussian process latent variable models and variational auto-encoders (VAEs) as special cases. VAEs are artificial neural networks (ANNs) that employ approximations to make computation tractable; however, current implementations lack adequate uncertainty quantification in estimating the parameters, predictive densities, and lower-dimensional subspace, and can be unstable and lack interpretability in practice. We attempt to solve these problems by deploying Markov chain Monte Carlo sampling algorithms (MCMC) for Bayesian inference in ANN models with latent variables. We address issues of identifiability by imposing constraints on the ANN parameters as well as by using anchor points. This is demonstrated on simulated and real data examples. We find that current MCMC sampling schemes face fundamental challenges in neural networks involving latent variables, motivating new research directions.
\end{abstract}


\section{Introduction}

Deep Learning (DL) has emerged as one of the most successful tools in image and signal processing. Excellent predictive performance in various high profile applications has lead to an enormous increase in focus on DL methods in a wide variety of fields, including artificial intelligence, astrophysics, chemistry and material science, healthcare, and manufacturing. The DL paradigm relies on multi-layer neural networks, which have been around for decades, with the recent success due mostly to advances in computing power, algorithms, and the availability of enormous datasets for training. This has been propelled by the rise of graphical processing units (GPUs). Many dedicated toolboxes have been developed and continue to be developed, including Apache MXNet, Caffe, the Microsoft Cognitive Toolkit, PyTorch, and TensorFlow. Many thousands of papers are published every year on deep learning.

The meteoric rise in the field of DL in the last decade started with a key paper of Geoffrey Hinton and collaborators showing that multi-layer convolutional neural networks could be tuned to imaging data to obtain a large improvement in classification performance \citep{krizhevsky2012imagenet}.  This stimulated a flood of researchers into the field, particularly in the imaging and signal processing communities, with a more recent spillover into statistics and other fields. The vast majority of the intellectual  energy in studying DL approaches has been focused on numerical optimization and engineering methods -- trying to modify the neural network architecture, tune hyper-parameters, and automate optimization algorithms to obtain good performance on test data in a variety of settings. 

\begin{figure}
\centering
\includegraphics[width=0.55\textwidth]{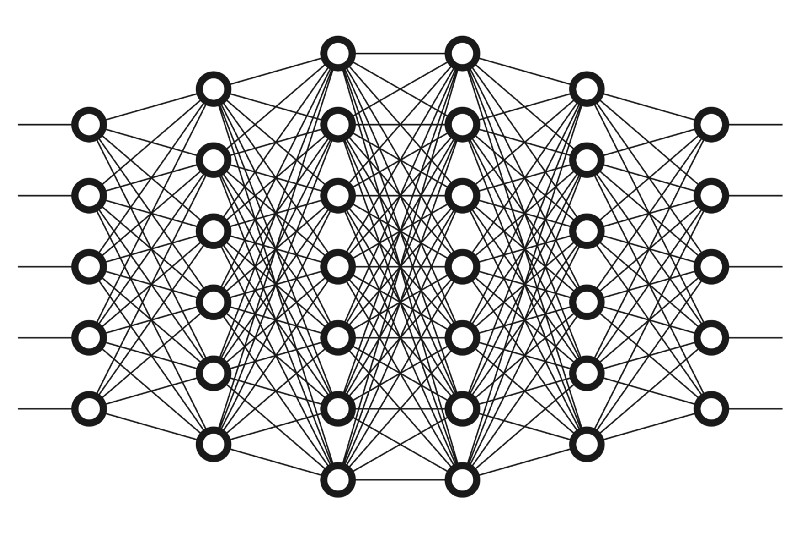} 
\caption{A neural network with four hidden layers. The left most layer is the input layer with five neurons, the right most layer is the output layer with five neurons, and the four hidden layers have six, seven, seven, and six neurons from left to right, respectively.}
\label{fig.nn}
\end{figure}

A neural network is a non-linear function $f$ consisting of an input layer, one or more hidden layers, and an output layer (\cref{fig.nn}). In addition, a non-linear \emph{activation function} is applied element-wise to each input of each layer.
Mathematically, this can be written as 
\begin{equation} \label{eq.nn.basic}
Y^i
=
f(X^i) 
=
W_4 \activ \left [ W_3 \activ \left \{W_2 \activ (W_1 X^i + b_1) + b_2 \right \} + b_3 \right ] + b_4 + \error^i,
\end{equation}
where $W_1,W_2,\dots$ are \emph{weights}, $b_1, b_2,\dots$ are \emph{biases}, $\activ : \RR \to \Omega \subseteq \RR$ is the non-linear activation function, $\error^i$ denotes an error term, and $i$ indexes the observations. Common activation functions include  sigmoid  $\activ(x) = \{1+\exp(-x)\}^{-1}$, the rectified linear unit (ReLU) $\activ(x) = x \, \indicator(x \geq 0)$, and the leaky ReLu $\activ(x) = 0.01 x \, \indicator(x<0) + x \, \indicator(x \geq 0)$, where $\indicator$ is the indicator function. 
The universal approximation theorem for neural networks \citep{cybenko1989approximation} guarantees that the function $f$ can be made arbitrarily flexible by adjusting the number of layers of the neural network and the number of hidden neurons per layer. 

In what follows, we let $\theta = (W_1, b_1, W_2, b_2, \dots)$ denote all the parameters of the neural network; this includes the weights and biases. Given training data $\{(X^i, Y^i)\}_{i=1}^N$, neural network training typically involves optimizing a loss function $L(\theta)$ as follows:
\begin{equation} \label{eq.nn.optim}
\theta_\star
=
\argmin_\theta 
\left \{
L(\theta) = 
\frac1N \sum_{i=1}^N \{ Y^i - f_\theta(X^i) \}^2 + p_0(\theta)
\right \}, 
\end{equation}
where $p_0(\theta)$ denotes a regularization term.
The optimization in \cref{eq.nn.optim} is usually achieved through a technique known as \emph{backpropagation} \citep{rumelhart1986learning}, which allows gradients with respect to $\theta$ of the loss function $L(\theta)$ to be calculated. The number of observations $N$ is usually very large and stochastic gradient-based algorithms are used for finding local optima of the non-convex loss function $L(\theta)$. Clever variants to usual stochastic gradient descent have been devised such as including momentum \citep{sutskever2013importance} and moment-based algorithms \citep{kingma2015adam}.
 
Dimensionality reduction is very important in making sense of the very high-dimensional and complex data routinely collected in 
modern scientific and technological applications. High-dimensional data tend to be highly structured and often can be accurately summarized with much lower-dimensional latent variables.  Linear dimensionality reduction methods produce a lower-dimensional linear mapping of the original high-dimensional dataset. This includes techniques such as principal component analysis, factor analysis, linear multidimensional scaling, Fisher's linear discriminant analysis, et cetera; we refer the reader to \cite{cunningham2015linear} for a detailed discussion. While linear methods have advantages in terms of interpretability and simplicity, they are limited in their ability to concisely represent complex data. This has motivated a rich literature on non-linear dimensionality reduction, including popular manifold learning algorithms 
Isomap \citep{tenenbaum2000global} and locally-linear embeddings \citep{roweis2000nonlinear}. These approaches learn a mapping from the high-dimensional observation space to a low-dimensional space that preserves certain desirable aspects of the original data such as pairwise distances.

An alternate generative approach for modelling high-dimensional data assumes that the data are generated by transforming a lower-dimensional latent variable $X^i$ to a higher-dimensional observation $Y^i$ through a function $g$ as $Y^i = g(X^i, \error^i)$, where $\error^i$ is a random error component.  For example, a linear latent factor model is obtained when $g(X^i,\error^i) = AX^i + \error^i$, with $\error^i \sim \N(\mu, \Sigma)$ for a factor loadings matrix $A$. More generally, the function $g$ can be a non-linear function of $X$; particular instances include Gaussian processes \citep{lawrence2004gaussian} and neural networks; we are interested in the latter approach in this text. Variational Bayes algorithms have been devised for neural networks with latent variables, and such methods are known as variational auto-encoders (VAEs; \citealp{kingma2014stochastic}). 

VAEs are  a popular generative modelling tool, which (in their most basic form) assume that model \eqref{eq.nn.basic} is of the form 
\begin{equation*} 
Y^i = f_\theta(X^i) + \error^i,
\quad X^i \sim g,\quad 
\error^i \sim \nu, 
\end{equation*}
where $f_\theta$ is a neural network (called the decoder) with $\theta$ denoting all the weights and biases in it, $\theta = (W_1, b_1, W_2, b_2, \dots)$.
The latent distribution $g$ is usually assumed to be $\N(0,I)$, although extensions ranging from mixture distributions \citep{dilokthanakul2016deep} to sequential models \citep{chung2015recurrent} to incorporating geometric aspects \citep{mathieu2019hierarchical} have been considered. VAEs have been used successfully in a wide range of contexts, including music generation \citep{roberts2017hierarchical}, image generation  \citep{dosovitskiy2016generating}, 
forecasting \citep{walker2016uncertain}, text generation \citep{semeniuta2017hybrid}, and clustering \citep{jiang2016variational}. We refer the reader to \cite{MAL-056} for an excellent introduction to VAEs.

VAEs employ a variational approximation to the distribution of $X^i \mid Y^i$. In particular, this is approximated by $\N\{\mu_\phi(Y^i), \Sigma_\phi(Y^i)\}$, where $\mu_\phi(\cdot)$ and $\Sigma_\phi(\cdot)$ are estimated through a neural network parameterised by $\phi$; this is known as the encoder. This is a restrictive assumption, as the actual distribution of $X \mid Y$ can be far from a Gaussian distribution. Moreover, a practical problem with VAEs is reproducibility, which is crucial in scientific applications. VAEs can be used as a dimension reduction technique to obtain lower dimensional representations $X^1, \dots, X^N$ of observations $Y^1, \dots, Y^N$, respectively. However, this is typically not reproducible: if the VAE is trained again to obtain $X^{1}_\prime, \dots, X^{N}_\prime$, these will typically be quite different from $X^1, \dots, X^N$. Variational approximations also lack theoretical guarantees in general and lead to inaccurate uncertainty quantification. Motivated by this, we adopt a fully Bayesian approach in this text. 

The rest of the text is organised as follows. We provide background on Bayesian parametric techniques for neural networks in \cref{sec.BNN.background}. \cref{sec.latent.bnn} is devoted to the main theme of this text -- Bayesian neural networks for dimensionality reduction; this includes techniques to reduce identifiability issues for such problems. \cref{sec.simulation} and \cref{sec.real.data} provide experimental assessment of our techniques based on simulated and real data, respectively.
Finally, \cref{sec.open.questions} concludes the text and discusses open questions and future research directions.

\section{Background on Bayesian neural networks}
\label{sec.BNN.background}

Variational approximations and Markov chain Monte Carlo algorithms provide alternative approaches for Bayesian inference in neural networks.  The former approach dominates the current literature, and MCMC implementations are relatively rare, likely due to the tendency of standard MCMC approaches to have poor performance. We provide a brief review of both strategies in this section.  

\subsection{Variational inference}
\label{sec.vb}

Variational Bayes methods enable approximate Bayesian inference
for challenging probability densities. The idea behind most  variational inference algorithms is to posit a parametric family of densities and then to find via optimization the member of the family that is closest to the target density. Closeness is evaluated using the Kullback-Leibler (KL) divergence or some distance in the space of probability densities. We refer the reader to \cite{blei2017} for an introduction to variational inference.

Designing effective variational approximations for Bayesian neural network (BNN) posteriors is a non-trivial task. For instance, it has been shown that more expressive variational families can yield significantly worse predictions in comparison to predictions obtained from less expressive variational families \citep{trippe2018}. Regularization techniques can improve the predictive generalization of variational inference in BNNs. This includes pruning \citep{graves2011} or an appropriate choice of priors, such as horseshoe shrinkage \citep{ghosh2018} or hierarchical priors \citep{wu2018}.
\cite{esmaeili2019, huang2019, titsias2019} developed variational families with hierarchical structure to approximate BNN posteriors. Such hierarchical variational models induce correlation among parameters to approximate the posterior covariance structure.
\cite{liu2016} introduced Stein variational gradient descent and applied it to BNNs. This is a variational inference framework that relies on a functional form of stochastic gradient descent to approximate the target density by iteratively transporting a set of particles. This links the derivative of KL divergence with Stein's identity and with the kernel Stein discrepancy. Further, \cite{shi2018} proposed a spectral Stein gradient estimator using Stein's identity, and \cite{sun2018} harnessed it to develop a functional form of variational inference for BNNs.

\subsection{Markov chain Monte Carlo}
\label{sec.bnn.mcmc}

Markov chain Monte Carlo (MCMC) methods tend to be slower to implement than variational algorithms, but have the advantage of (conceptually) providing posterior summaries that converge to the true values as the number of Monte Carlo samples increases.  However, MCMC algorithms for BNNs face several challenges \citep{papamarkou2019challenges}. Firstly, the huge size of both the parameter space and input data space of BNNs can render MCMC algorithms computationally infeasible. Secondly, weight symmetries in ANNs create identifiability issues and yield multimodal weight posteriors \citep{pourzanjani2017}: MCMC algorithms waste computational time exploring posterior modes associated with equivalent solutions in the output space \citep{nalisnick2018}. Third, BNNs are complex non-linear hierarchical models, and such models are known to cause problems with MCMC \citep{zhang2014}. Fourth, it is not clear how to choose good default priors for BNNs \citep{lee2004,lee2005,vladimirova2019understanding}. Several types of priors have been studied, including truncated flat priors \citep{lee2005}, restricted flat priors \citep{lee2003}, Jeffreys priors \citep{lee2007}, smoothing priors \citep{freitas1999}, Laplace priors \citep{williams1995} and approximate reference priors \citep{nalisnick2018}, but it remains to find good default priors.
 
These challenges have led to limited literature on MCMC algorithms for BNNs. Sequential  Monte  Carlo  and  reversible jump MCMC were applied to multilayer perceptrons (MLPs) and radial basis function networks as early as two decades back \citep{andrieu1999sequential, freitas1999, andrieu2000, freitas2001}, but have not seen much development since. Geometric Langevin Monte Carlo and population MCMC (power posterior sampling) were also applied to MLPs \citep{papamarkou2019challenges}. To the best of our knowledge, there are not any instances of exact MCMC algorithms for convolutional neural networks (CNNs) in the literature.

To scale MCMC methods with big data, a log-likelihood can be evaluated on a subset (minibatch) of the data rather than on the entire dataset. The notion of minibatch was employed by \cite{welling2011bayesian} to develop a stochastic gradient Monte Carlo algorithm. \cite{chen2014} introduced stochastic gradient Hamiltonian Monte Carlo and applied it to BNNs. \cite{gong2018} proposed a stochastic gradient MCMC scheme that generalizes Hamiltonian dynamics with state-dependent drift and diffusion, and demonstrated the performance of this scheme on CNNs and on recurrent neural networks.

\section{Latent factor models through neural networks} 
\label{sec.latent.bnn}

\subsection{General model} 

Consider the situation where we have observed data $Y^1, \dots, Y^N \in \RR^p$, and assume that these lie near a lower-dimensional manifold. We consider latent factor models of the form
\begin{align} \label{eq.gen_model}
\begin{aligned}
X^i 
& \iid 
g(\cdot),
\quad 
Y^i \mid X^i 
\ind 
\N \{ f_\theta(X^i), \tau^2 \}, \quad i = 1, \dots, N,
\end{aligned}
\end{align}
with latent factors $X^i \in \RR^q$ for $q<p$ and $i = 1, \dots, N$. In addition, we let $f_\theta$ be a neural network with $\theta$ denoting all the weights and biases in the network. Let $\bY^N = (Y^1, \dots, Y^N)$ and $\bX^N = (X^1, \dots, X^N)$ for compact notation. We place a prior $p_0(\theta)$ on $\theta$ and are interested in the posterior $p(\theta, \bX^N \mid \bY^N)$.

For simplicity, consider a neural network with a single hidden layer with $h$ neurons. This is also known as a multilayer perceptron (MLP). Following \cref{eq.nn.basic}, the function $f_\theta : \RR^q \mapsto \RR^p$ can be written as 
\begin{align} \label{eq.1layer_nn}
f_\theta(X^i) 
= 
W_2 \activ (W_1 X^i + b_1) + b_2 \in \RR^p,
\end{align}
for some $h\times q$ matrix $W_1$ and
$p\times h$ matrix $W_2$, leading to the log-likelihood
\begin{align} \label{eq.log_ll}
\begin{aligned}
\ell(\bY^N \mid \bX^N, \theta)
& \propto 
- \frac1{\tau^2} \sum_{i=1}^N \{ f_\theta(X^i) - Y^i \}^2 
\\
& =
- \frac1{\tau^2} \sum_{i=1}^N \{ W_2 \activ (W_1 X^i + b_1) + b_2 - Y^i \}^2. 
\end{aligned}
\end{align}

In general, the model specified by \cref{eq.gen_model} is not identifiable;
different values $\bX^N$ and $\theta$ give the same likelihood of $\bY^N$.
Instead of focusing on identifiability of $\bX^N$ and $\theta$, we focus on identifiability of pairwise distances between the latent $X^i$s. This is akin to dimension reduction techniques which aim to preserve pairwise distances between the latent variables, such as Isomap \citep{tenenbaum2000global}, locally-linear embeddings \citep{roweis2000nonlinear}, and t-distributed stochastic neighbor embedding \citep{maaten2008visualizing}.

Let $X^i = (X^i_1, \dots, X^i_q)^T \in \RR^q$, and $W_1 \in \RR^{h \times q}$ be written in a column-wise fashion as $W_1 = (w_{11}, \dots, w_{1q})$, where $w_{1j}$ is a $h$-dimensional column vector for $j = 1, \dots, q$. Then we have $W_1 X^i = \sum_{j=1}^q w_{1j} X^i_j$, which is invariant up to a permutation of $\{1, \dots, q\}$: if $\{\permutation(1), \dots, \permutation(N)\}$ denotes a permutation of $\{1, \dots, N\}$, then $\sum_{j=1}^q w_{1j} X^i_j = \sum_{j=1}^q w_{1 \permutation(j)} X^i_{\permutation(j)}$. This does not pose a problem since we are interested only in pairwise distances between the $X^i$s. In the next three Sections \ref{sec.identifiability}-\ref{sec.anchor+constr}, we consider more carefully the identifiability issue.

\subsection{Identifiability} 
\label{sec.identifiability}

Formally, having an identifiable model would mean that 
\begin{align}
\ell (\bY^N \mid \bX^N, \theta, \tau^2)
& = 
\ell \{ \bY^N \mid \bX^N_\prime, \theta_\prime, (\tau_\prime)^2 \} 
~ \forall ~ \bY^N \implies
\label{eq.idef_equality}
\\
\bX^N = \bX^N_\prime,  
~~ 
\theta 
& = 
\theta_\prime,
~~
\tau^2 = (\tau_\prime)^2. \nonumber 
\end{align}
Equation \eqref{eq.idef_equality} implies 
\begin{align*}
\frac1{\tau^2} \sum_{i=1}^N \{ f_\theta(X^i) - Y^i \}^2 
& =  
\frac1{(\tau_\prime)^2} \sum_{i=1}^N \{ f_{\theta_\prime}(X^i_\prime) - Y^i \}^2 \implies
\\
\frac1{\tau^2} \sum_{i=1}^N \left [ \{ f_\theta(X^i) \}^2 + (Y^i)^2 - 2 f_\theta(X^i) Y^i \right ]
& = 
\frac1{(\tau_\prime)^2} \sum_{i=1}^N \left [ \{ f_{\theta_\prime}(X^i_\prime) \}^2 + (Y^i)^2 - 2 f_{\theta_\prime}(X^i_\prime) Y^i \right ],
\end{align*}
which in turn implies 
\begin{equation} \label{eq.idef}
\tau^2 
= 
(\tau_\prime)^2,
\quad
f_\theta(X^i) 
= 
f_{\theta_\prime}(X^i_\prime),
\quad \text{and} ~~
\sum_{i=1}^N \{ f_\theta(X^i) \}^2 
= 
\sum_{i=1}^N  \{ f_{\theta_\prime}(X^i_\prime) \}^2.
\end{equation}
The third equality of \cref{eq.idef} follows from the second, and so we focus only on the third equality. We now isolate some factors that cause identifiability issues. 

Since we are only interested in pairwise distances among the $X^i$s, location shifts do not matter: $\dist(X^i, X^j) = \dist(X^i+c, X^j+c)$ for any $c \in \RR^q$, where $\dist$ denotes the Euclidean distance. Moreover, the model given by \cref{eq.1layer_nn} is not invariant to scale changes in the columns of $W_1$ as $\sum_{j=1}^q w_{1j} X^i_j = \sum_{j=1}^q c_j w_{1j} X^i_j/c_j$ for any $c_1, \dots, c_q \in \RR$. We address this by imposing a norm constraint on the columns of $W_1$ as follows:
\begin{equation} \label{eq.constrain_weights_1}
\|w_{1j}\|^2 = \sum_{k=1}^h w_{1jk}^2 = 1, \quad j = 1, \dots, q, 
\end{equation}
where $\| \cdot \|$ denotes the Euclidean norm. This has the effect of fixing the scale of the latent variables. Typically, constraints are included in Bayesian inferences by choosing priors to have constrained support.  However, such direct approaches can be computationally challenging, motivating recent proposals including constraint relaxation \citep{duan2020bayesian} and posterior projections  \citep{patra2020constrained}.  
Fortunately, we can straightforwardly impose \cref{eq.constrain_weights_1} by re-parameterizing \cref{eq.1layer_nn} as 
\begin{equation} \label{eq.scale}
f_\theta(X^i) 
= 
W_2 \activ \left \{ \begin{pmatrix} w_{11}/\|w_{11}\| & \cdots & w_{1q}/\|w_{1q}\| \end{pmatrix} X^i + b_1 \right \} + b_2. 
\end{equation}

\begin{remark}
The leaky ReLU activation function is linear, therefore the rows of $W_2$ have to be taken to have unit length according to $\|w_{2 \cdot i}\|^2 = \sum_{j=1}^p w_{2ij}^2 = 1$, $i=1, \dots, h$, if using it as the activation function. In this text, we consider the hyperbolic tangent activation function $\activ(x) = \{\exp(x)-\exp(-x)\}/\{\exp(x)+\exp(-x)\}$, and so we do not need to impose this constraint on the rows of $W_2$.
\end{remark}

\subsection{Anchor points} 
\label{sec.anchor.points}

The idea of \emph{anchor points} has been used in the context of Gaussian mixture models to address the label switching problem  \citep{kunkel2018anchored}. The main idea behind this is to ``anchor'' a certain number of latent variables, while conducting a full Bayesian inference for the remaining latent variables. This can be viewed as placing a highly informative point mass prior for the anchored variables. We use the idea of anchor points to improve identifiability for our model.

We fix $N_\rf$ anchor points out of a total $N$ latent variables as follows. We start by assuming that we know the ``true'' latent representations of $\{Y^i\}_{i \in \A_\rf}$ to be $\{X^i\}_{i \in \A_\rf}$, respectively, where $\A_\rf \subset \{1, \dots, N\}$ denotes indices of anchor points, with $|A_\rf| = N_\rf$. Let $\A_\rf^c = \{1, \dots, N\} \setminus \A_\rf$ denote indices of points other than the anchor points.
We define the \emph{anchored posterior} as
\begin{equation} \label{eq.anchored.posterior}
\pi(\{ X^i \}_{i \in \A_\rf^c}, \, \theta \mid \bY^N) 
\propto 
\prod_{i \in \A_\rf^c} \{ p(Y^i \mid X^i, \theta) \, p_0(X^i) \} 
\prod_{i \in \A_\rf} p(Y^i \mid X^i, \theta) 
\times 
p_0(\theta). 
\end{equation}
The posterior for the non-anchored latent variables quantifies uncertainty in the latent representation. Recall that our focus is is Bayesian inference on the pairwise distances, and not the latent variables directly. In our experiments in Sections \ref{sec.simulation} and \ref{sec.real.data}, we choose $N_\rf$ to be less than $6\%$ of $N$, which means that only $0.36\%$ of pairwise distances are pre-specified.

The question now is how to obtain the anchored $\{X^i\}_{i \in \A_\rf}$. Any one of a number of nonlinear dimension reduction techniques can be used for this. In this text, we use locally-linear embeddings \citep{roweis2000nonlinear}. This approximately preserves pairwise distances among the lower-dimensional representation of the observations. More precisely, we consider the locally-linear embeddings of the observations $\{Y^i\}_{i=1}^N$. Let these lower-dimensional points be $\{Y^i_\lle \}_{i=1}^N$. We let $\A_\rf$ be chosen randomly without replacement from $\{1, \dots, N\}$.

\subsection{Combining parameter constraints with anchor points} 
\label{sec.anchor+constr}

As discussed in \cref{sec.anchor+constr}, we also impose constraint \eqref{eq.constrain_weights_1} on the columns of $W_1$. The experiments of Section \ref{sec.simulation} demonstrate empirically that anchors points together with parameter constraints improve the mixing properties of MCMC sampling. In this case, fixing the reference points to be $\{Y^i_\lle \}_{i \in \A_\rf}$ will cause a scaling issue. To see this, note that if all the latent variables $\{X^i\}_{i=1}^N$ were unrestricted, imposing constraint \eqref{eq.constrain_weights_1} leaves the model unchanged as discussed in the previous \cref{sec.identifiability}. However, since we have some fixed anchor points $\{Y^i_\lle\}_{i \in \A_\rf}$, this changes the model as we are now forcing the neural network to map \emph{fixed} $\{Y^i_\lle \}_{i \in \A_\rf}$ to $\{Y^i\}_{i \in \A_\rf}$ while imposing constraint \eqref{eq.constrain_weights_1}, and mapping unrestricted $\{X^i\}_{i \in \A_\rf^c}$ to $\{Y^i\}_{i \in \A_\rf^c}$. Having a constraint on the weights reduces the ability of a neural network to map $\{Y^i_\lle \}_{i \in \A_\rf}$ to $\{Y^i\}_{i \in \A_\rf}$.

To address this issue, we train an initial neural network using stochastic gradient descent with $\{Y^i_\lle\}_{i \in \A_\rf}$ as input and $\{Y^i\}_{i \in \A_\rf}$ as output. Let the weights and biases of the trained neural network be $W_1^\optim, \, b_1^\optim$, etc. We then multiply $\{Y^i_\lle \}_{i \in \A_\rf}$ by the norms of the columns of $W_1^\optim$ to obtain the anchored points $\{X^i\}_{i \in \A_\rf}$; that is, let $X^i_j = Y^i_{\lle,j} \|w_{1j}^\optim\|$, $j = 1, \dots, q$, $i \in \A_\rf$, where $X^i = (X^i_1, \dots, X^i_q)$, $Y^i_\lle = (Y^i_{\lle,1}, \dots, Y^i_{\lle,q})$, and $W_1^\optim = (w_{11}^\optim, \dots, w_{1q}^\optim)$ is written in a column-wise fashion as in \cref{eq.scale}. This leads to the anchor points that we use and is summarized in \cref{alg.anchor}.

\begin{algorithm}[t]
\caption{Obtaining anchor points while imposing constraint \eqref{eq.constrain_weights_1}}
\label{alg.anchor}
\textbf{Input:} Observations $Y^1,\dots,Y^N$. 
\begin{algorithmic}[1] 

\STATE 
Choose $\A_\rf$ randomly without replacement from $\{1, \dots, N\}$.

\STATE 
Let $\{Y^i_\lle\}_{i \in \A_\rf} = \mathrm{locally ~ linear ~ embedding}(\{Y^i\}_{i \in \A_\rf})$.

\STATE 
Let $W_1^\optim, \, b_1^\optim, \ldots = \argmin_{W_1,b_1,\ldots} \sum_{i\in\A_\rf} \{ W_2 \activ (W_1 Y^i_\lle + b_1) + b_2 - Y^i \}^2$.

\STATE 
For $i \in \A_\rf$, let $X^i_j = Y^i_{\lle,j} \|w_{1j}^\optim\|$, $j = 1, \dots, q$.

\end{algorithmic}

\textbf{Output:} Anchor points $\{(X^i, Y^i)\}_{i \in \A_\rf}$. 
\end{algorithm}

\section{Simulation study for hyper-sphere} 
\label{sec.simulation}

\subsection{Model and sampling details}
\label{sec.sphere.model.details}

\begin{figure}[H]
\centering
\includegraphics[width=0.55\textwidth]{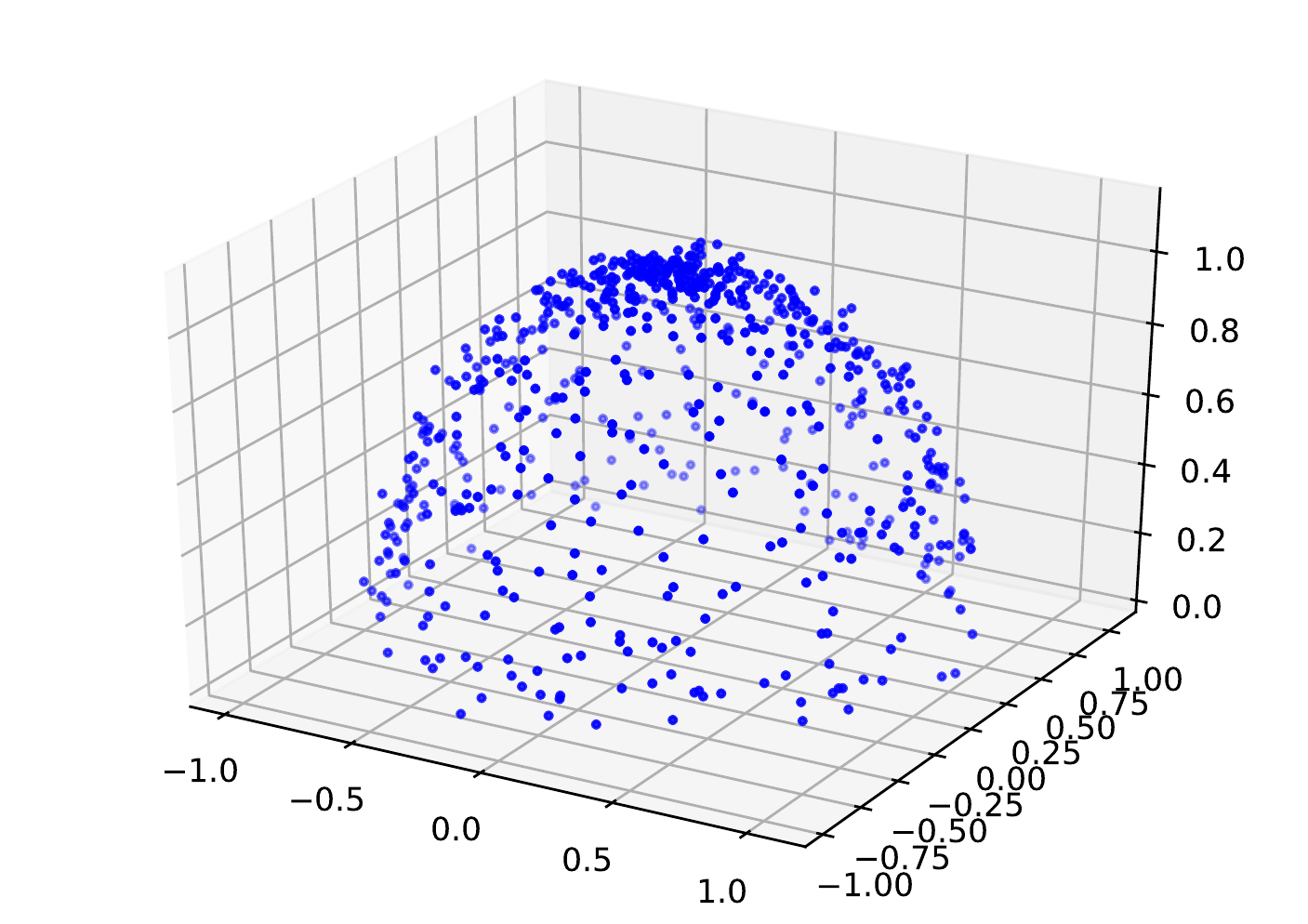} 
\caption{Noisy observations on hypersphere.}
\label{fig.sphere}
\end{figure}

We conduct a simulation study in which we have $N=640$ three-dimensional noisy observations lying on a hypersphere. The observations are plotted in \cref{fig.sphere}. The dimension of the latent manifold is two in this case. 

We use a multilayer perceptron with a single hidden layer with $h=10$ neurons and the hyperbolic tangent activation function. We place independent $\N(0,\sigma^2)$ priors on each component of $\theta$ (that is, on each component of $W_1, b_1$, et cetera). We place a $\hcauchy(5)$ prior on $\tau^2$ and a $\hcauchy(5)$ hyperprior on $\sigma^2$.  

As discussed in \cref{sec.BNN.background}, sampling from the posterior of the parameters of a neural network is a highly challenging problem in itself, and this is only exacerbated by the fact that we place a distribution on the input (latent) variables as well. For this reason, we do not focus on designing sampling algorithms in this text as this is a daunting task; instead, we use Hamiltonian Monte Carlo (HMC; \citealp{duane1987hybrid}). In particular, we consider the no-U-turn sampler \citep{hoffman2014no} as implemented in Stan \citep{gelman2015stan} as this provides a popular and automated approach for choosing the hyperparameters of HMC. The initial value of $\theta$ for the Markov chain is set randomly using Stan's \texttt{init="random"} command.

\subsection{Results}

\begin{figure}
\centering
\includegraphics[width=0.7\textwidth]{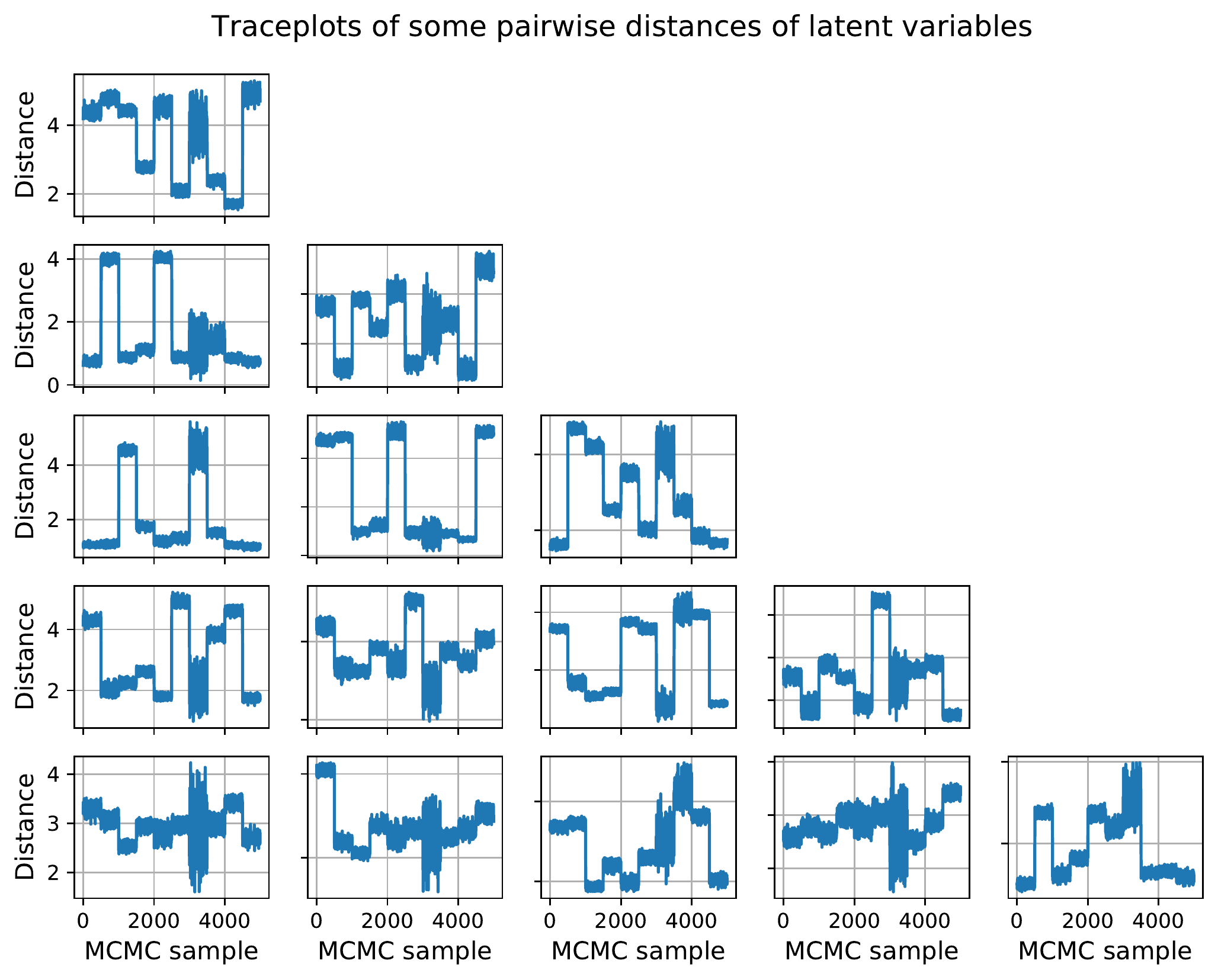} 
\caption{Traceplots of posterior samples of pairwise distances for randomly chosen latent variables when imposing constraint \eqref{eq.constrain_weights_1} on the weights $W_1$ and without anchor points for the hypersphere example.}
\label{fig.sphere_pwdist_constrainedW1_norefpoints}
\end{figure}

\begin{figure}
\centering
\includegraphics[width=0.7\textwidth]{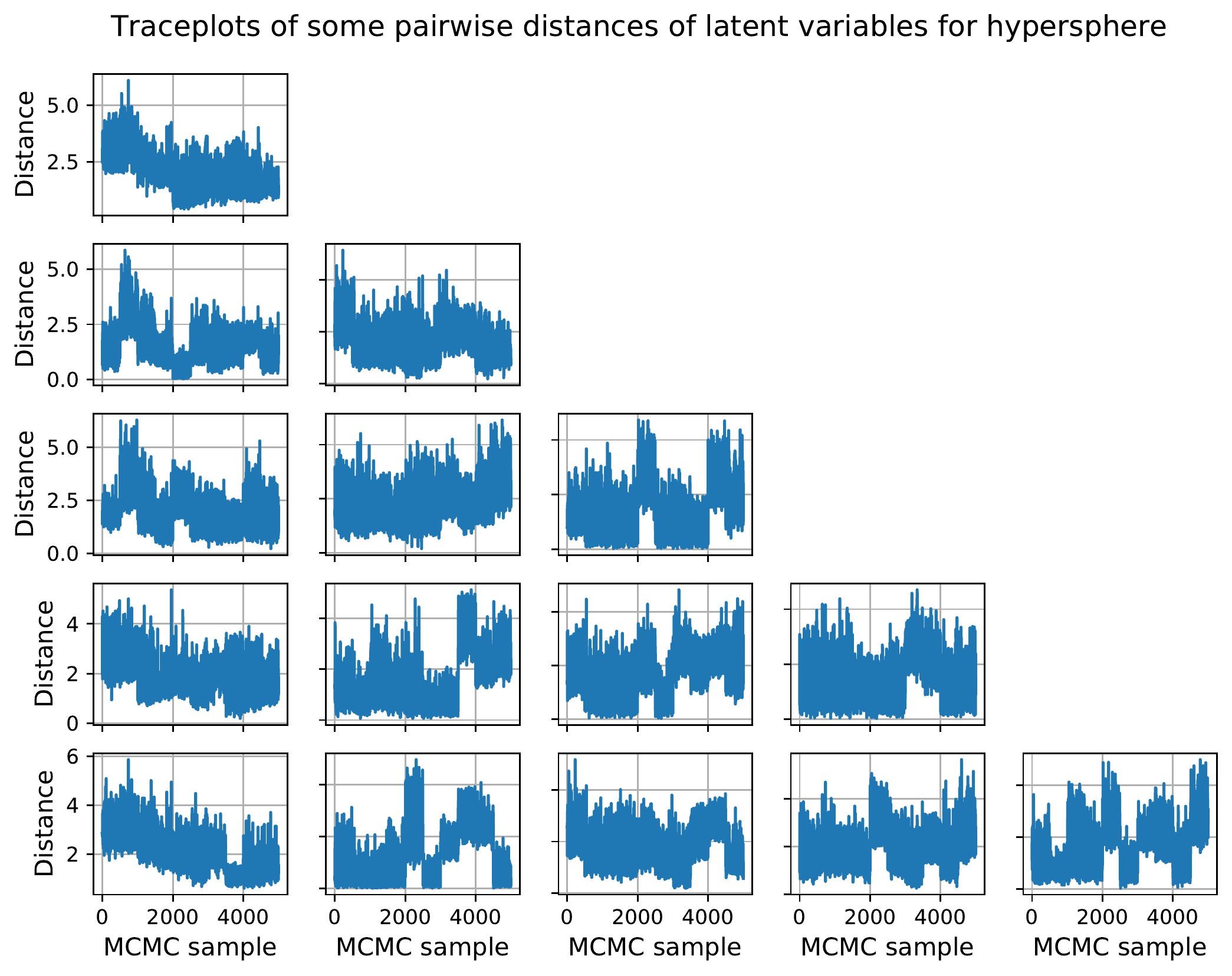} 
\caption{Traceplots of posterior samples of pairwise distances for randomly chosen latent variables without imposing constraints on the weights $W_1$ and with around $6\%$ anchor points for the hypersphere example.}
\label{fig.sphere_pwdist_unconstrainedW1_refpoints}
\end{figure}

\begin{figure}
\centering
\includegraphics[width=0.7\textwidth]{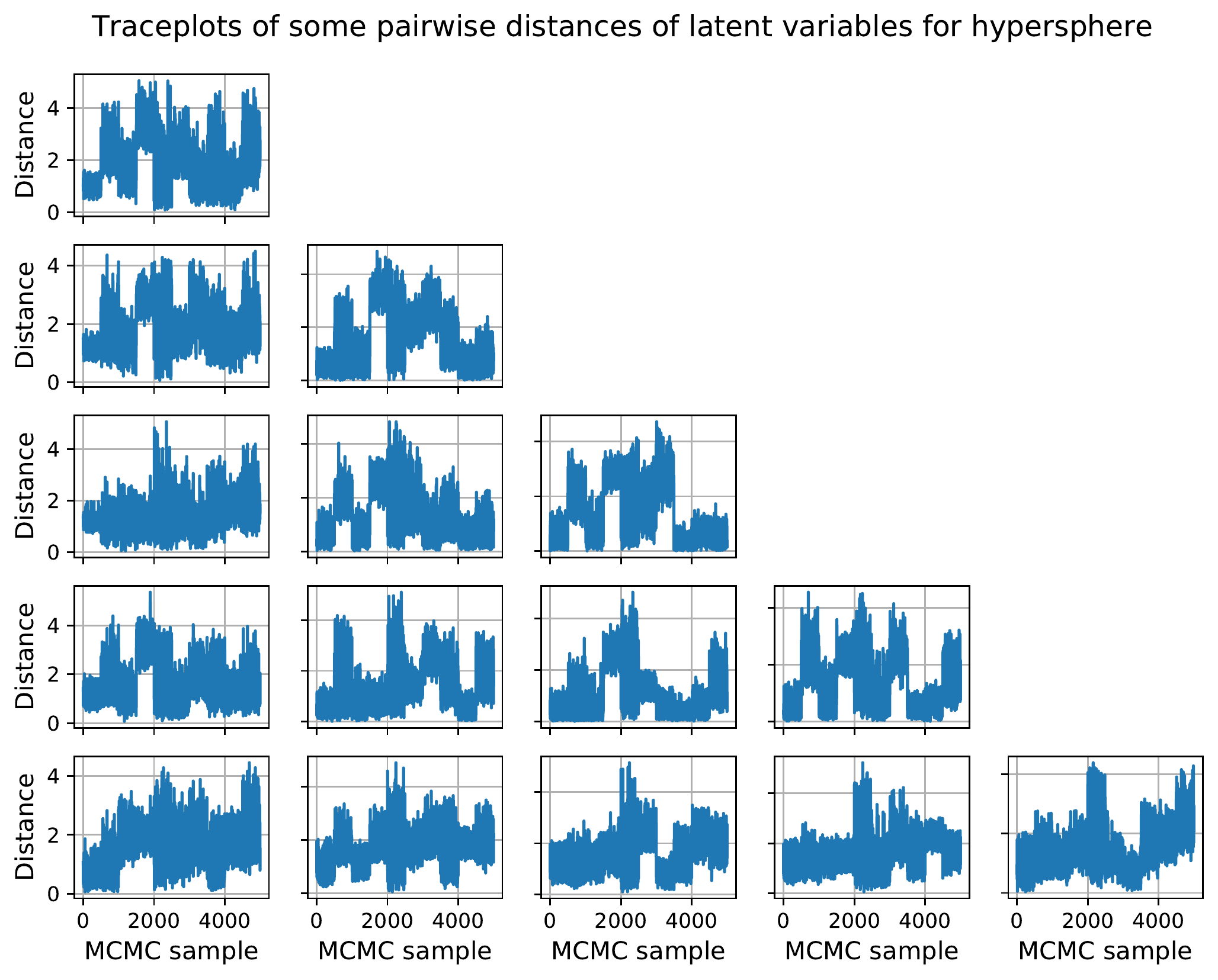} 
\caption{Traceplots of posterior samples of pairwise distances for randomly chosen latent variables without imposing constraints on the weights $W_1$ and with almost $20\%$ anchor points for the hypersphere example.}
\label{fig.sphere_pwdist_unconstrainedW1_more_refpoints}
\end{figure}

We use traceplots of the posterior samples of pairwise distances as a proxy for their identifiability. Badly mixing traceplots may indicate that the model suffers from identifiability issues.
In a first experiment, we consider only constraint \eqref{eq.constrain_weights_1} on the weights $W_1$ without anchor points. We randomly pick some pairs of latent variables and display the traceplots of pairwise distances in \cref{fig.sphere_pwdist_constrainedW1_norefpoints}; these figures show very poor mixing/convergence consistent with a lack of identifiability. In a second experiment, we choose $N_\rf = 40$ anchor points and do not impose constraint \eqref{eq.constrain_weights_1} on the weights $W_1$. The anchored points constitute about $6\%$ of all latent variables. The resulting posterior samples of pairwise distances of some randomly chosen pairs of latent variables is plotted in \cref{fig.sphere_pwdist_unconstrainedW1_refpoints}. We observe that while the traceplots indicate better mixing in comparison to \cref{fig.sphere_pwdist_constrainedW1_norefpoints},
mixing remains poor. We increase the number of anchor points to $N_\rf=120$ and display the traceplots of the pairwise distances in \cref{fig.sphere_pwdist_unconstrainedW1_more_refpoints}, which still exhibit slow mixing. This demonstrates that using even a relatively large number of anchor points (almost $20\%$ of the observations) is still not sufficient to ensure good mixing for the Markov chains of pairwise distances.

\begin{figure}
\centering
\includegraphics[width=0.7\textwidth]{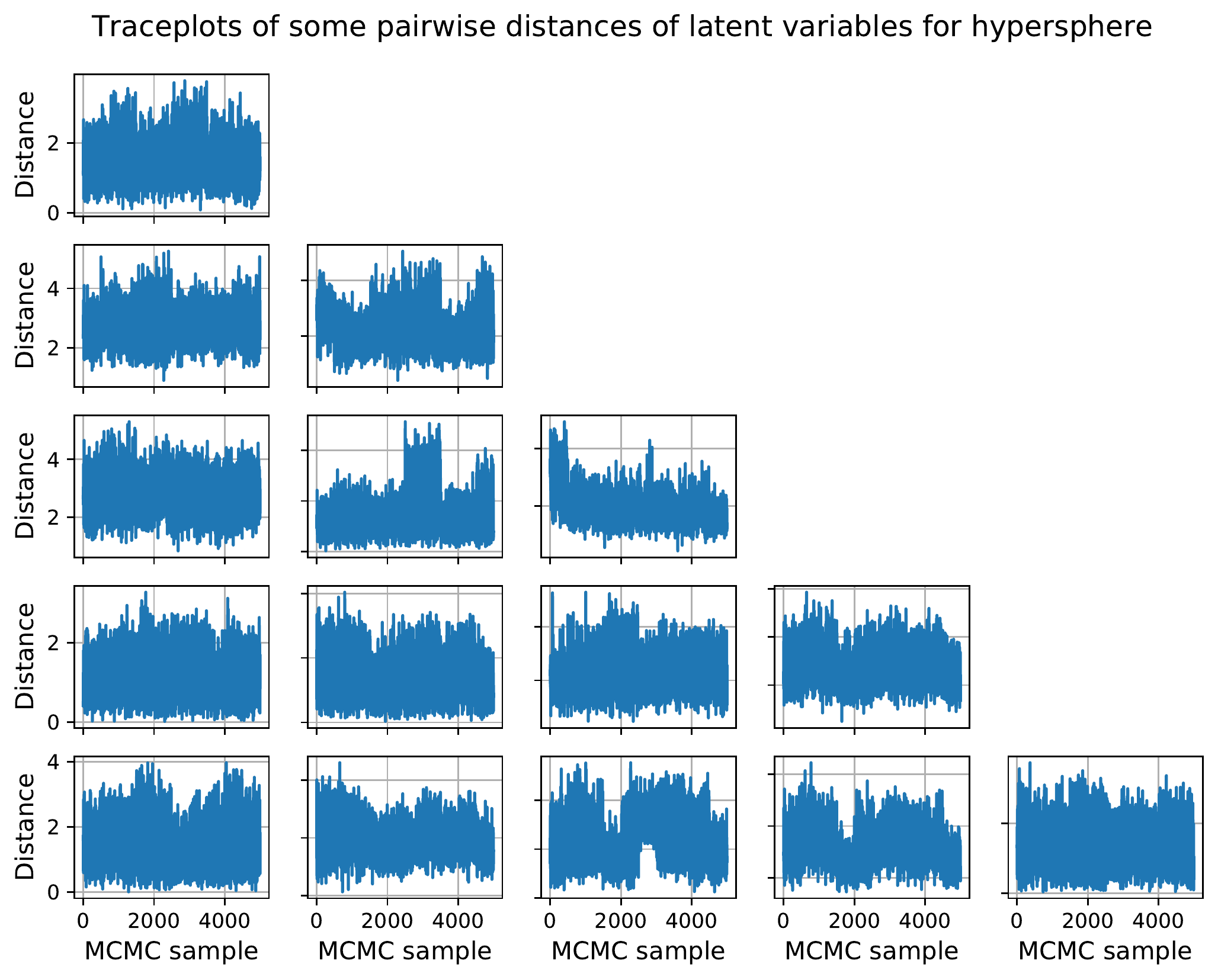} 
\caption{Traceplots of posterior samples of pairwise distances for randomly chosen latent variables when imposing constraint \eqref{eq.constrain_weights_1} on the weights $W_1$ and with around $6\%$ anchor points for the hypersphere example.}
\label{fig.sphere_pwdist_constrainedW1_refpoints}
\end{figure}

Finally, we consider both $N_\rf = 40$ anchor points as well as constraint \eqref{eq.constrain_weights_1} on the weights $W_1$; in this case, the anchor points are obtained as described in \cref{alg.anchor}. The resulting pairwise distances are plotted in \cref{fig.sphere_pwdist_constrainedW1_refpoints}, where we observe that the traceplots of pairwise distances attain significantly improved mixing (though there is still clear room for improvement). This provides evidence that the identifiability issue has been addressed, and using the posterior samples of such pairwise distances allows us to infer structure in the data, as will be done in the real data examples of the following \cref{sec.real.data}.

\begin{figure}
\centering
\begin{minipage}{0.49\textwidth}
\centering
\includegraphics[width=0.85\textwidth]{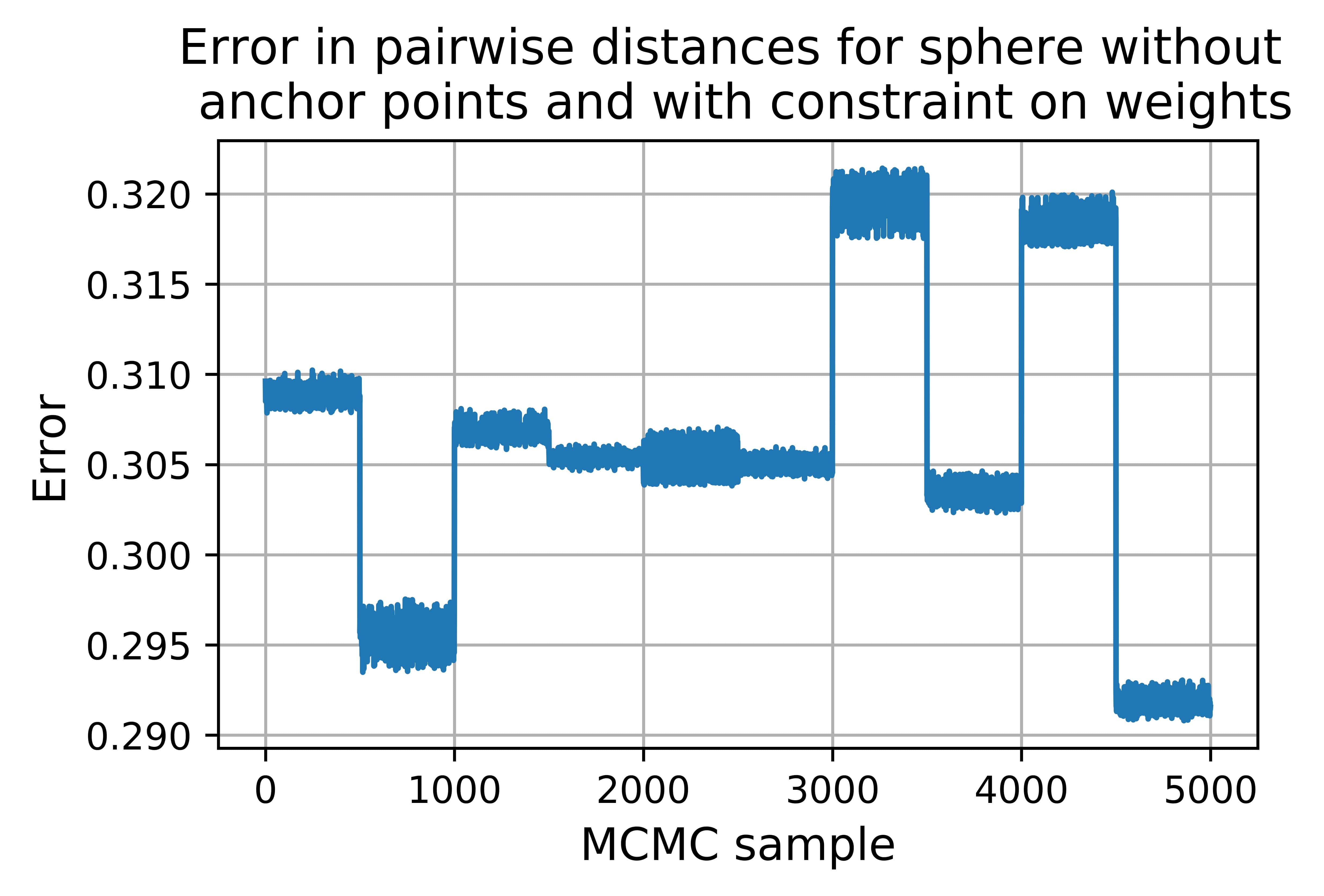}
\end{minipage}\hfill
\begin{minipage}{0.49\textwidth}
\centering
\includegraphics[width=0.9\textwidth]{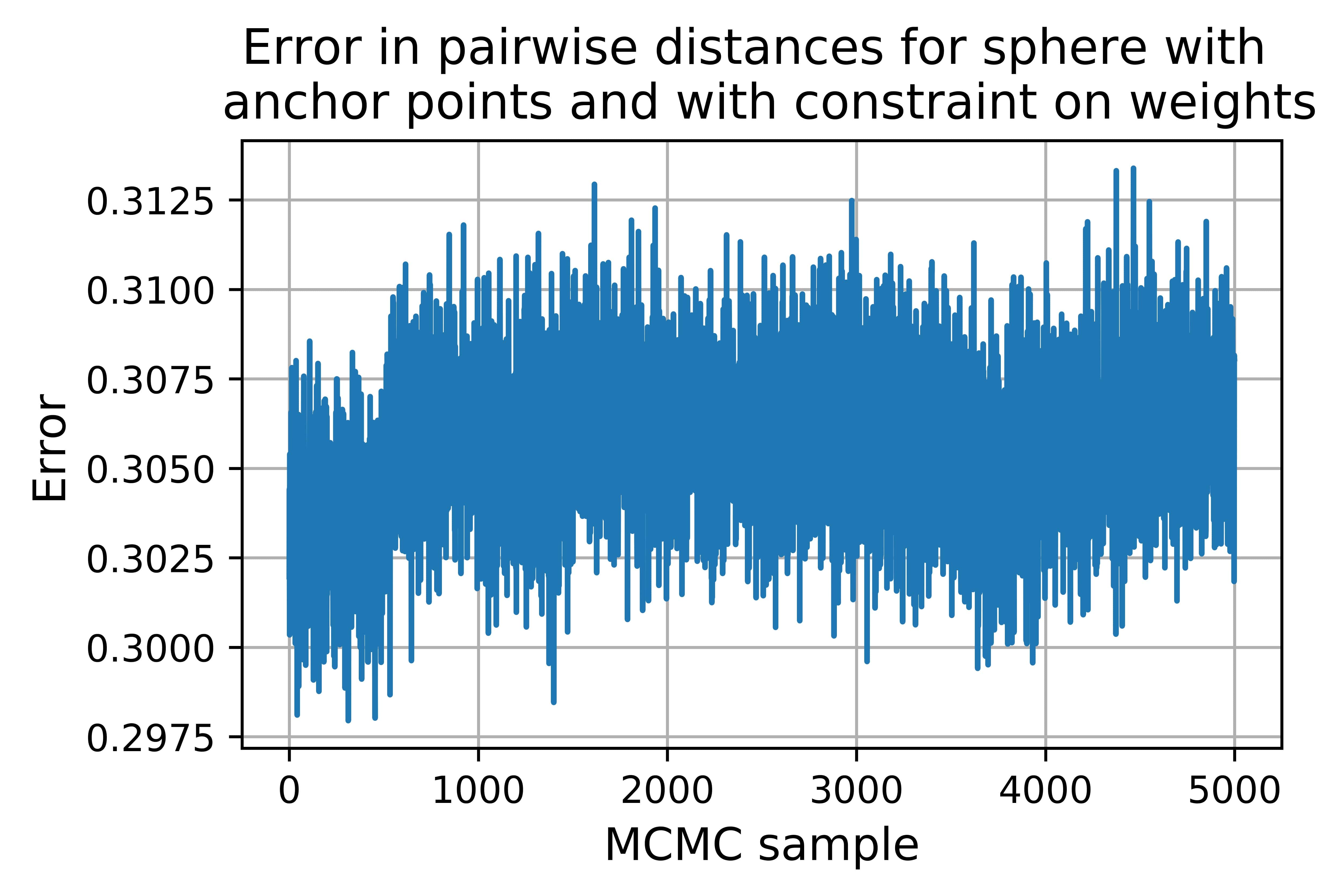}
\end{minipage}\hfill
\caption{Errors in estimating the pairwise distances for the hypersphere example.}
\label{fig.sphere.pwdist_error}
\end{figure}

We also compare the pairwise distances for the posterior samples with the pairwise distances of the true latent variables; this ground truth is known since we have simulated it. Let $D$ be the $N \times N$ matrix of the true pairwise distances, and let $D^{(k)}$ be the $N \times N$ matrix of the pairwise distances of the $k$th posterior sample; here $k=1,\dots,K$ indexes the posterior samples. We plot the error $(1/N) \|D^{(k)} - D\| = (1/N) [\sum_{i=1}^N \sum_{j=1}^N \{D^{(k)ij} - D^{ij}\}^2]^{1/2}$ for $k=1,\dots,K$ with constraint \eqref{eq.constrain_weights_1} and without and with anchor points, respectively, in \cref{fig.sphere.pwdist_error}. 
We have used ten chains in Stan.
We note that the pairwise distances are of a similar order, which is expected as Stan concentrates posterior samples in regions where the target density is high and we have fixed the scale of the latent representation by imposing constraint \eqref{eq.constrain_weights_1}. However, we note that when we do not use anchor points, the errors in pairwise distances are all clearly different for the ten chains (indeed, most of them are non-overlapping), which suggest identifiability issues: there are different ways of having a lower-dimensional latent representation.

\section{Real data applications}
\label{sec.real.data}

All datasets were obtained from the UCI machine learning repository \citep{Dua:2019}. For each application in this section, the prior for $\theta$ and initial values of the Markov chains are the same as described in \cref{sec.sphere.model.details}.

\subsection{E. coli data} 
\label{sec.ecoli}

We consider data on Escherichia coli (E. coli) proteins \citep{horton1996probabilistic}. Each observation consists of a classification label for the protein (with eight classes in total) and has seven attributes; we refer the reader to \cite{horton1996probabilistic} for details. The number of observations is $N=336$. We model the seven-dimensional attributes as lying on a one-dimensional latent manifold, and we use a multilayer perceptron with one hidden layer with five neurons. 

\begin{figure}
\centering
\includegraphics[width=0.7\textwidth]{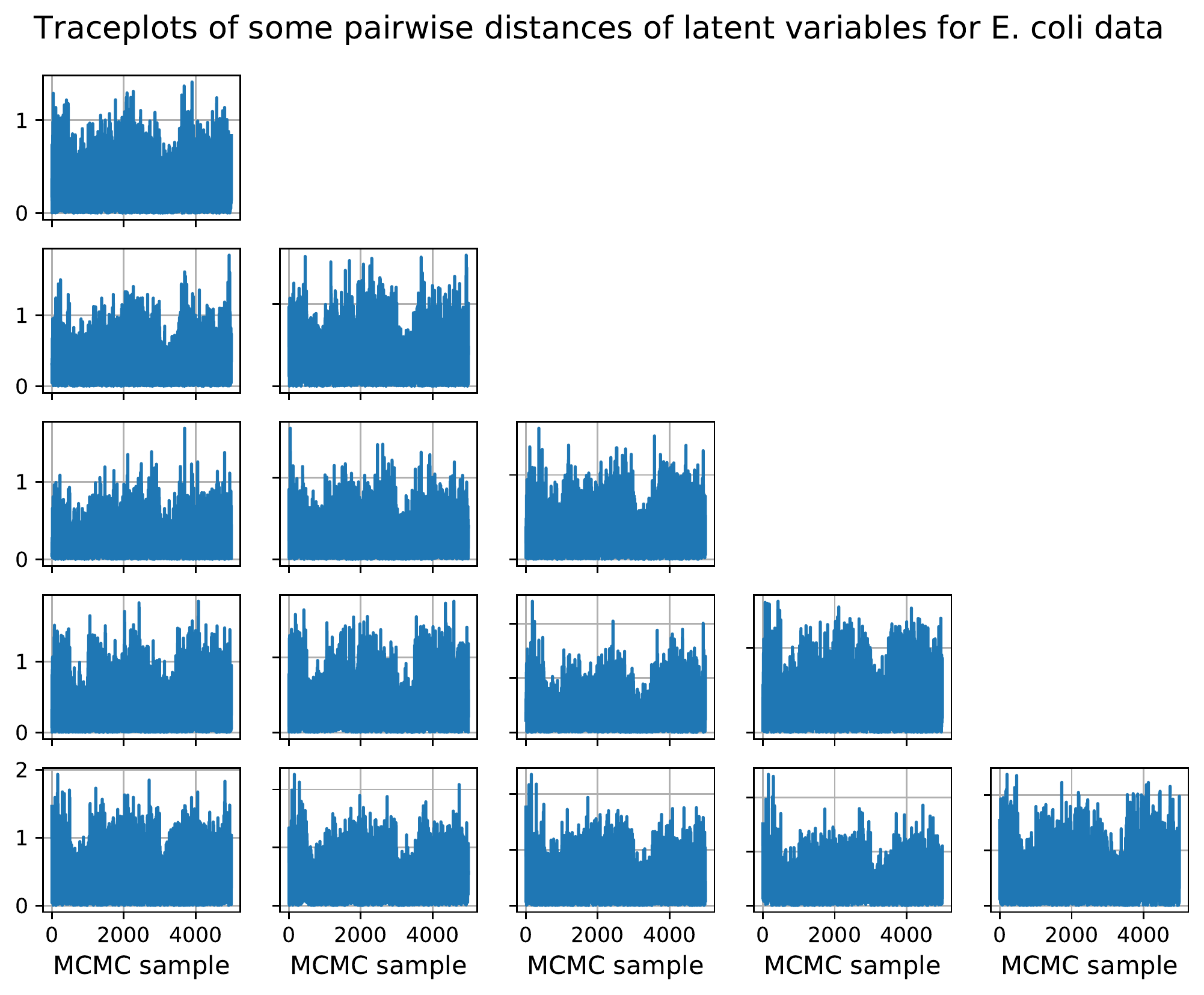} 
\caption{Traceplots of posterior samples of pairwise distances for randomly chosen latent variables when imposing constraint \eqref{eq.constrain_weights_1} on the weights $W_1$ and with anchor points for the E. coli dataset.}
\label{fig.ecoli_pwdist_constrainedW1_refpoints}
\end{figure}

In the previous Section \ref{sec.simulation}, we observed that imposing constraint \eqref{eq.constrain_weights_1} on the weights $W_1$ and tying down around $6\%$ of the observations as anchor points lead to adequate MCMC mixing and good performance in estimating the pairwise distances.
We therefore impose the constraint and consider $20$ anchor points as described in \cref{sec.anchor.points}. We plot traceplots of posterior samples of pairwise distances for some randomly chosen pairs of latent variables in \cref{fig.ecoli_pwdist_constrainedW1_refpoints} and note that they exhibit relatively good mixing (compared to analyses that do not include constraints -- improvements on the HMC algorithm being used may lead to further gains).

\begin{figure}
\centering
\begin{minipage}{0.49\textwidth}
\includegraphics[width=0.95\textwidth]{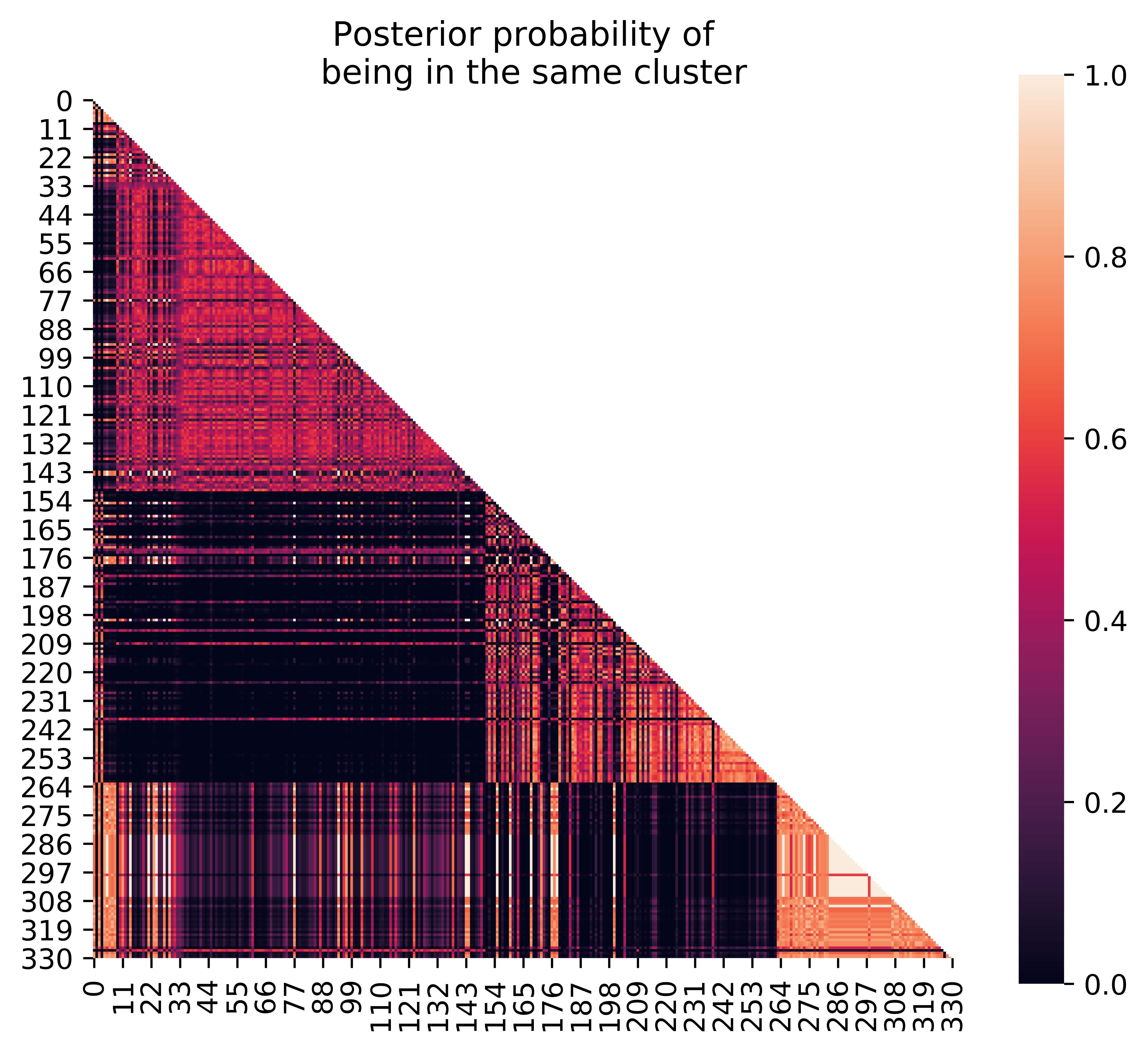}
\end{minipage}\hfill
\begin{minipage}{0.49\textwidth}
\includegraphics[width=1.03\textwidth]{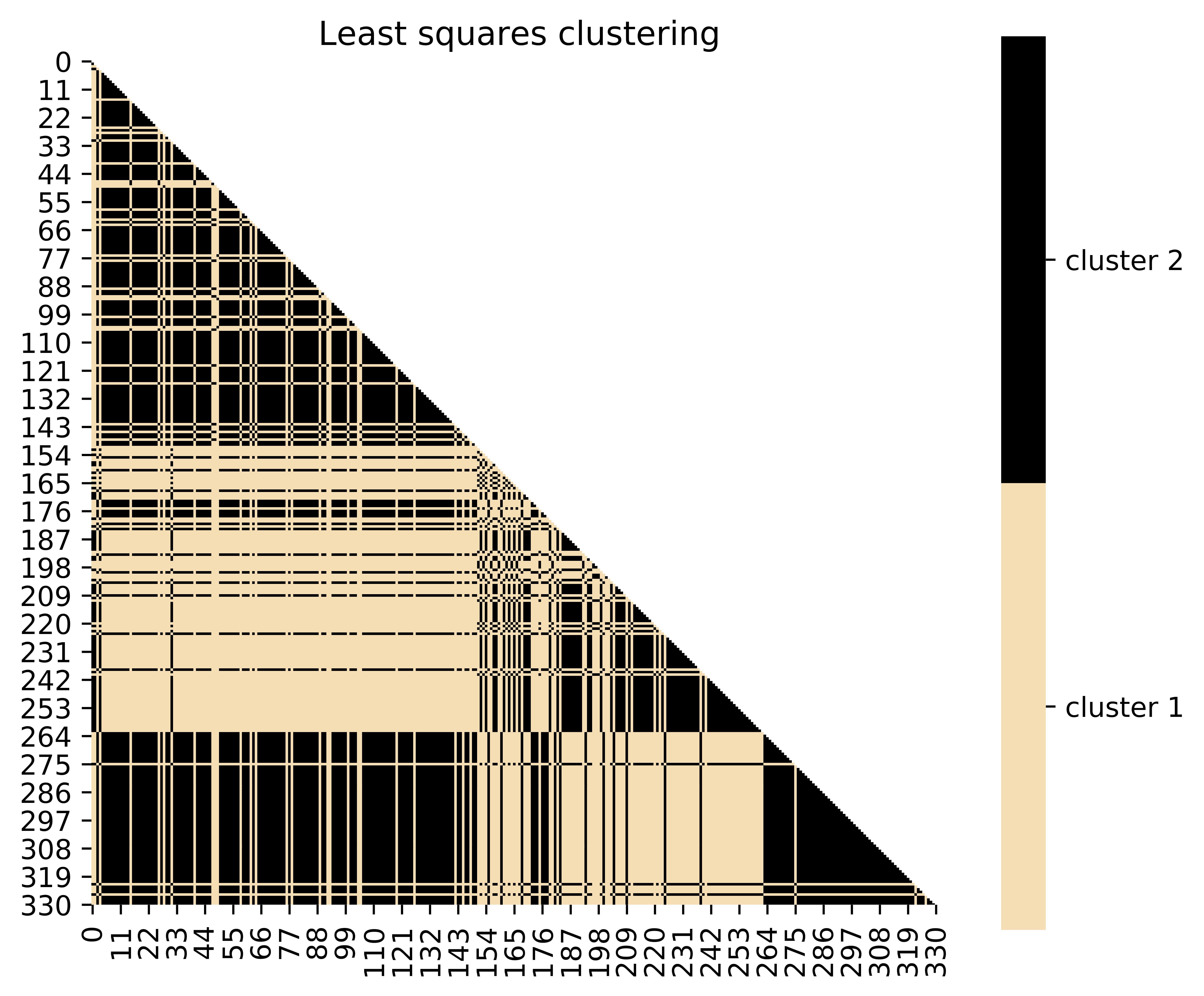}
\end{minipage}\hfill
\caption{Posterior probability of pairs of observations belonging to the same cluster (left figure)
and least squares clustering of \cite{dahl2006model} (right figure)
for the E. coli dataset.}
\label{fig.ecoli_clustering}
\end{figure}

We use the posterior samples of the pairwise distances to uncover some structure in the dataset. In particular, we use spectral clustering \citep{spielman1996spectral}, as implemented in Python's \texttt{sklearn.cluster} command, to obtain a clustering of the latent variables for every posterior sample; we refer the reader to \cite{von2007tutorial} for a detailed introduction to spectral clustering. Based on this clustering, we form an $N \times N$ clustering matrix $C^{(k)}$ for each posterior sample $\bX^{N(k)} = (X^{1(k)}, \dots, X^{N(k)})$, where 
\begin{equation*}
C^{(k)}_{ij}
=
\begin{cases}
1  & \quad \text{if } X^{i(k)} ~ \text{and} ~ X^{j(k)} ~ \text{are in the same cluster}, \\
0  & \quad \text{otherwise};
\end{cases}
\end{equation*}
here $k=1,\dots,K$ indexes the posterior samples. 
The posterior mean $(1/K) \sum_{k=1}^K C_{ij}^{(k)}$ estimates the probability that observations $i$ and $j$ belong to the same cluster. These are plotted in the left panel of \cref{fig.ecoli_clustering}, where the observations are sorted by their true class labels. Moreover, using these probabilities as a pairwise affinity matrix and applying the least squares clustering method of 
 \cite{dahl2006model} leads to an estimated clustering of the data samples. This is plotted in the right panel of \cref{fig.ecoli_clustering}. Some structure is visible here, with a few classes being distinguishable.

\subsection{User knowledge data} 
\label{sec.user.knowledge}

\begin{figure}
\centering
\includegraphics[width=0.7\textwidth]{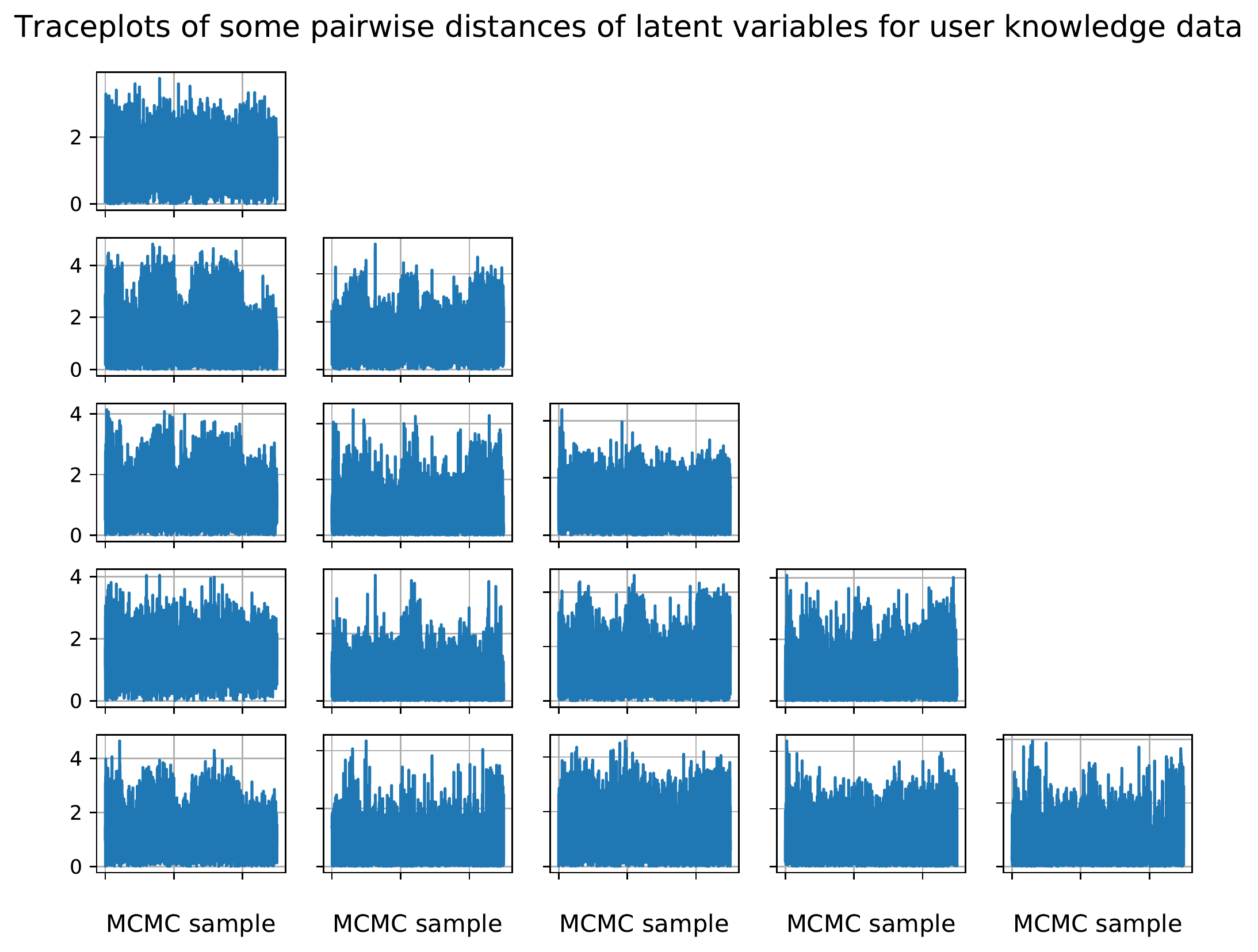} 
\caption{Traceplots of posterior samples of pairwise distances for randomly chosen latent variables when imposing constraint \eqref{eq.constrain_weights_1} on the weights $W_1$ and with anchor points for user knowledge dataset.}
\label{fig.user_knowledge_pwdist_constrainedW1_refpoints}
\end{figure}

\begin{figure}
    \centering
    \begin{minipage}{0.49\textwidth}
        \includegraphics[width=0.95\textwidth]{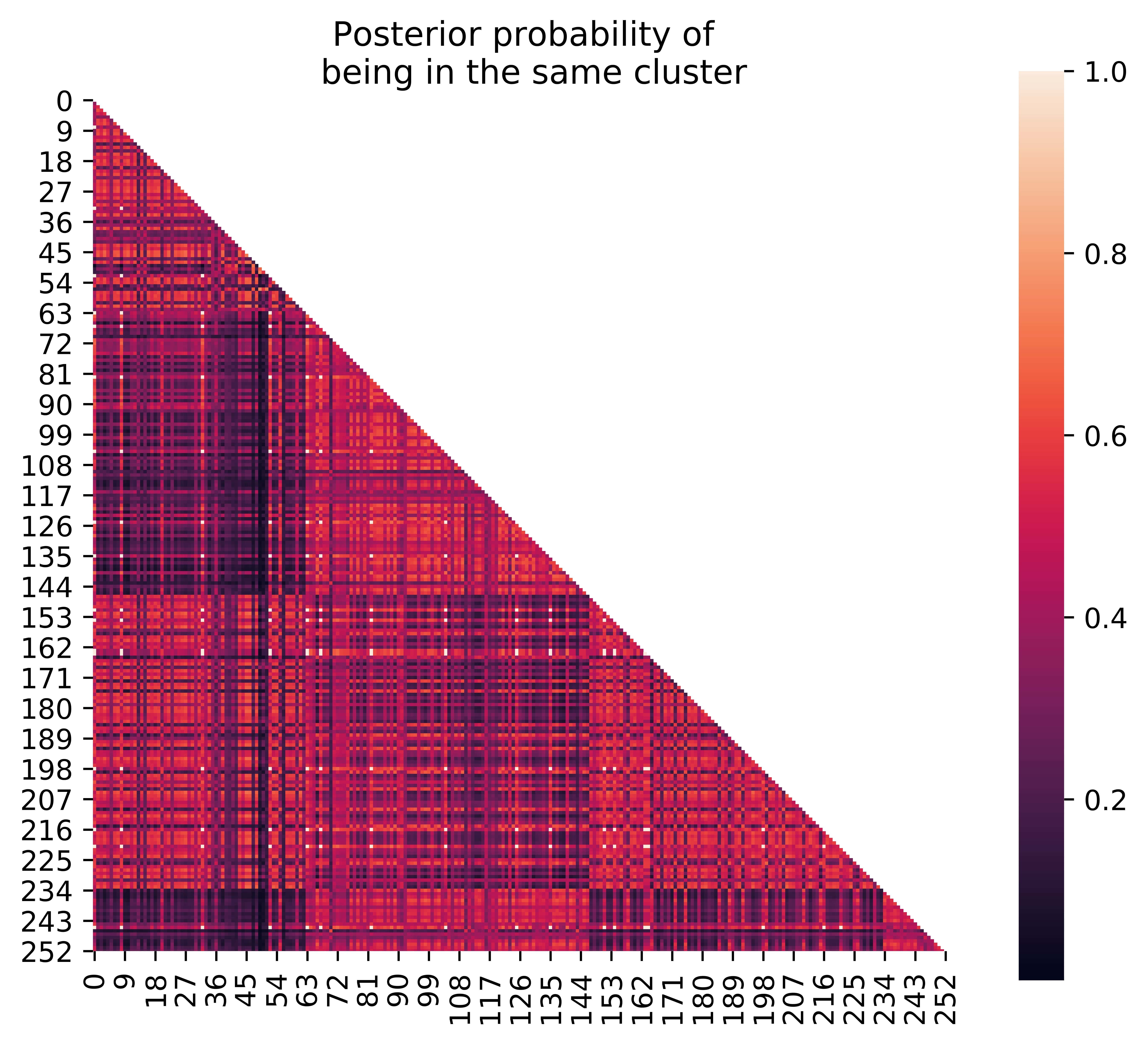}
    \end{minipage}\hfill
    \begin{minipage}{0.49\textwidth}
        \includegraphics[width=1.03\textwidth]{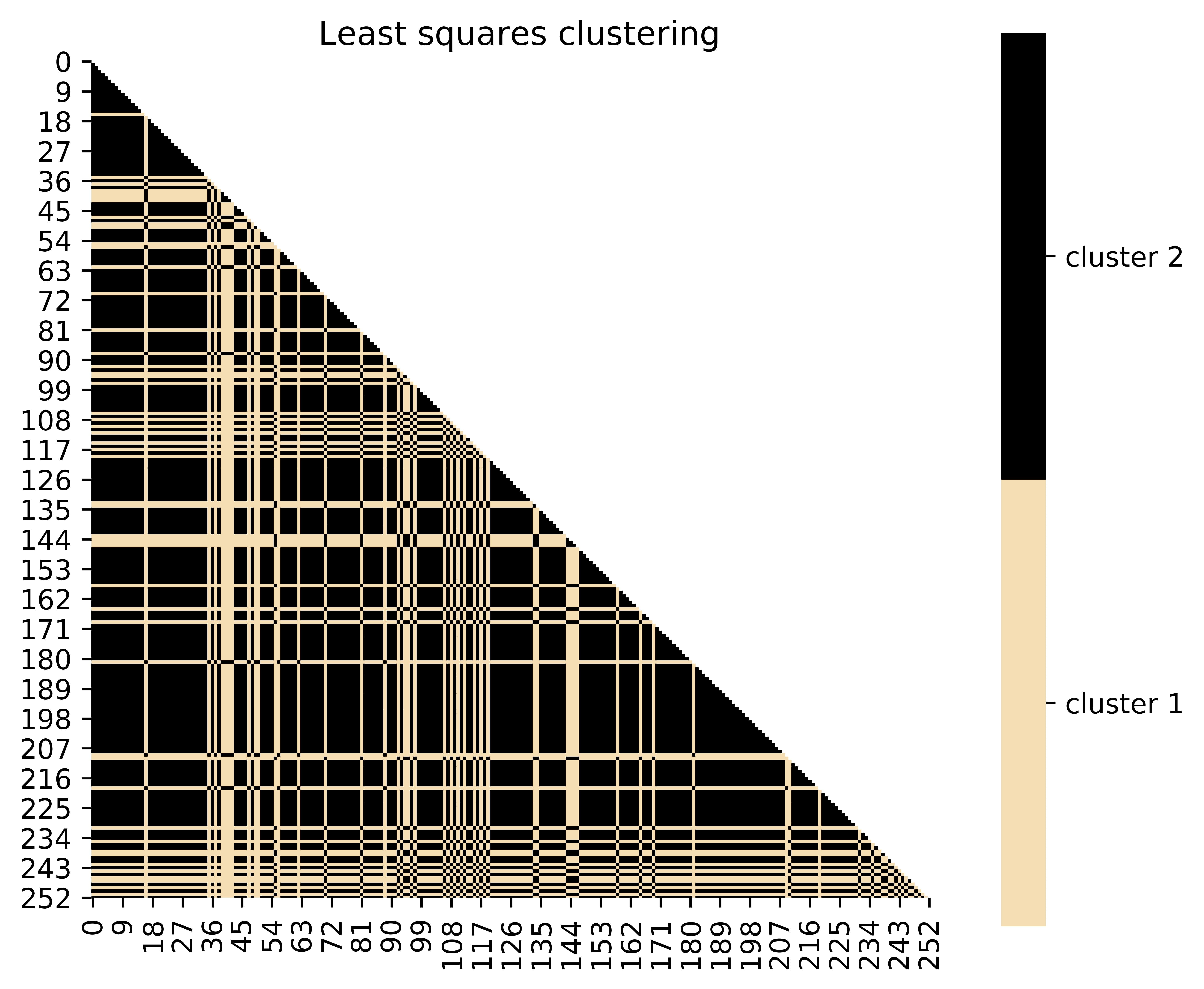}
    \end{minipage}\hfill
    \caption{Posterior probability of pairs of observations belonging to the same cluster (left figure) and least squares clustering of \cite{dahl2006model} (right figure) for the user knowledge dataset.}
    \label{fig.user_knowledge_clustering}
\end{figure}

In a second application, we consider the user knowledge data \citep{kahraman2013development}. This consists of five dimensional attributes, including study time for goal object materials and for related objects materials. The number of observations is $N=257$, with each observation being additionally classified into one of four classes indicating the knowledge level of user.
We model the attributes as lying on a one-dimensional latent manifold. We use a multilayer perceptron with one hidden layer consisting of $h=10$ neurons. As in the E. coli data (\cref{sec.ecoli}), we consider constraint \eqref{eq.constrain_weights_1} and additionally tie down $N_\rf = 15$ anchor points. Traceplots of posterior samples of pairwise distances of some  some randomly chosen pairs of latent variables, plotted in \cref{fig.user_knowledge_pwdist_constrainedW1_refpoints}, demonstrate relatively good mixing (again with substantial room for improvement).
As in \cref{sec.ecoli}, we also plot the posterior probabilities of pairs of observations being in the same cluster and well as the least squares cluster in \cref{fig.user_knowledge_clustering}, where the observations are sorted by their true class labels. This figure indicates that the uncertainty in clustering using the lower-dimensional representation of the attributes is high.

\subsection{Banknote data}
\label{sec.banknote}

\begin{figure}
\centering
\includegraphics[width=0.7\textwidth]{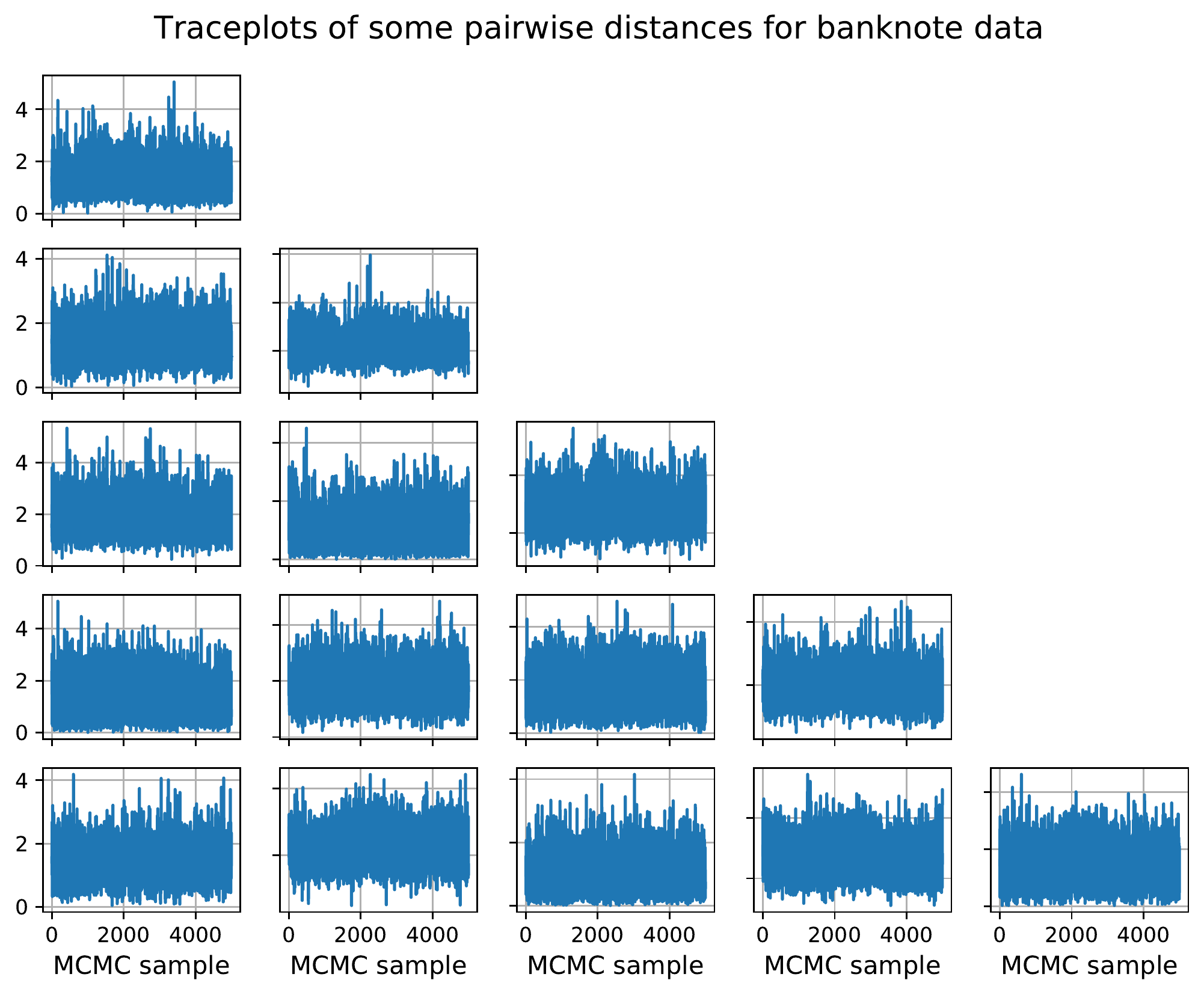} 
\caption{Traceplots of posterior samples of pairwise distances for randomly chosen latent variables when imposing constraint \eqref{eq.constrain_weights_1} on the weights $W_1$ and with anchor points for the banknote dataset.}
\label{fig.banknote_pwdist_constrainedW1_refpoints}
\end{figure}

We consider data consisting of four-dimensional attributes which were extracted from images that were taken from genuine and forged banknote-like specimens. 
Wavelet transform tools were used to extract four features from images. 
Each of $N=1371$ observations is further classified into genuine and fake. We model the four-dimensional attributes as lying on a two-dimensional latent manifold, and we use a multilayer perceptron with one hidden layer with $h=20$ neurons. We impose constraint \eqref{eq.constrain_weights_1} and consider eighty anchor points as described in \cref{sec.anchor.points}. We plot some pairwise distances of posterior samples of the latent variables in \cref{fig.banknote_pwdist_constrainedW1_refpoints}, where we note that the Markov chains of pairwise distances are mixing adequately. As in \cref{sec.ecoli}, we also plot the posterior probabilities of pairs of observations being in the same cluster as well as the least squares cluster in \cref{fig.user_knowledge_clustering}, where the observations are sorted by their true class labels. The uncertainty in clustering using the lower-dimensional representation is very high in this case.

\begin{figure}
    \centering
    \begin{minipage}{0.49\textwidth}
        \includegraphics[width=0.95\textwidth]{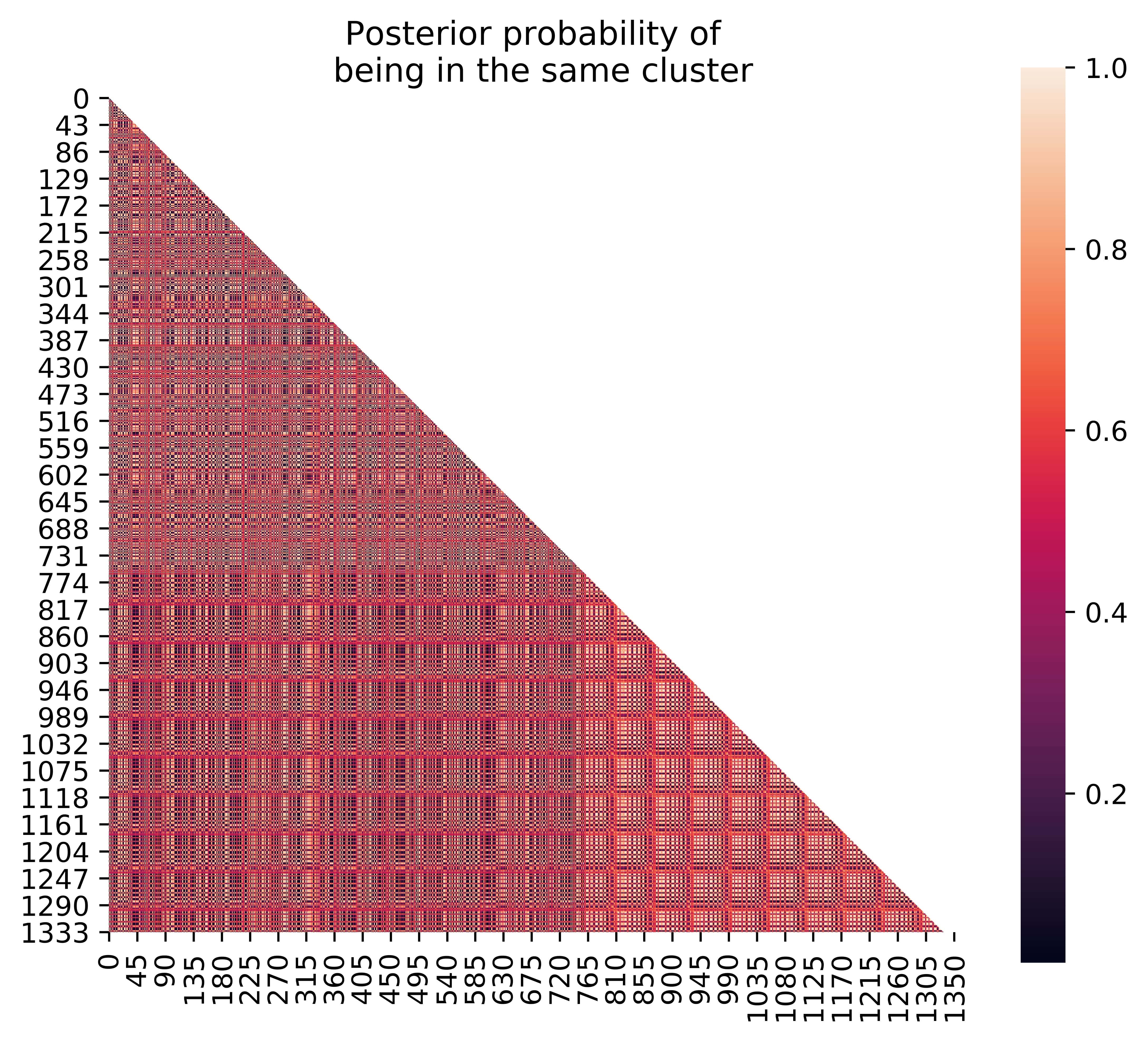}
    \end{minipage}\hfill
    \begin{minipage}{0.49\textwidth}
        \includegraphics[width=1.03\textwidth]{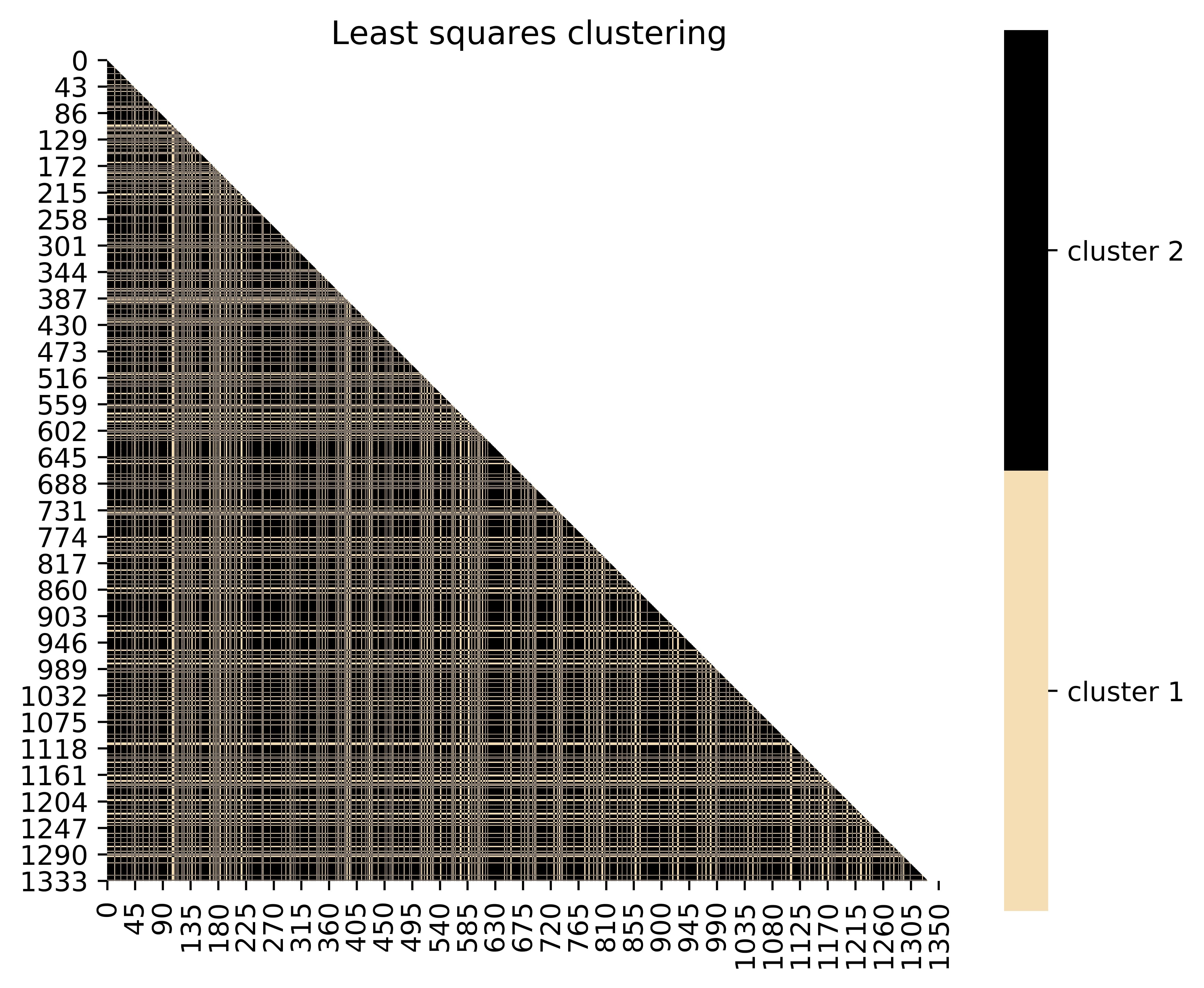}
    \end{minipage}\hfill
    \caption{Posterior probability of pairs of observations belonging to the same cluster (left figure) and least squares clustering of \cite{dahl2006model} (right figure) for the banknote dataset.}
    \label{fig.banknote_clustering}
\end{figure}

\section{Open questions and ongoing research}
\label{sec.open.questions}

Efficient Bayesian inference for the posterior of the parameters of a neural network remains an open question.
Neural networks have routinely been used in many tech industries, and a recent paper \citep{wenzel2020good} has made the point that as of mid-2020, there are no publicized deployments of Bayesian neural networks in industry. They further argue that sampling the posterior predictive yields worse predictions as compared to simpler methods such as stochastic gradient descent. As we discussed in \cref{sec.bnn.mcmc}, choosing priors for the parameters of a neural network remains an open question as well. 

Bayesian inference for neural networks based on MCMC methods has not flourished due to various challenges posed by neural networks, such as the high number of their parameters, their hierarchical model structure and non-trivial likelihood landscape, and their likelihood function complexity and associated identifiability and label switching issues. In this text, we demonstrate that parameter constraints can alleviate identifiability and anchor points can mitigate the label switching problem for neural networks, thus yielding MCMC sampling with much better (but still not ideal) mixing for statistics of interest, such as pairwise distances for latent factors. A potentially promising avenue for future MCMC research for neural networks may thus focus on estimating lower dimensional summaries of the parameters by inducing parameter constraints and by exploiting re-labelling algorithms. Such summaries may involve pairwise distances of latent variables or the posterior covariance structure of the parameters.  Alternatively, new classes of MCMC algorithms that can more adeptly explore and mix across the complex posterior landscape need to be developed.  

\appendix

\section*{Acknowledgements}

DS acknowledges support from grant DMS-1638521 from SAMSI.
DD acknowledges National Institutes of Health (NIH) grant R01ES027498 and Office of Naval Research (ONR) grant N00014-17-1-2844.

This manuscript has been authored by UT-Battelle, LLC, under contract DE-AC05-00OR22725 with
the US Department of Energy (DOE). The US government retains and the publisher, by accepting
the article for publication, acknowledges that the US government retains a nonexclusive, paid-up,
irrevocable, worldwide license to publish or reproduce the published form of this manuscript, or allow
others to do so, for US government purposes. DOE will provide public access to these results of
federally sponsored research in accordance with the DOE Public Access Plan (http://energy.gov/down-
loads/doe-public-access-plan). This research was sponsored by the Laboratory Directed Research
and Development Program of Oak Ridge National Laboratory, managed by UT-Battelle, LLC, for the
US Department of Energy under contract DE-AC05-00OR22725.

\bibliographystyle{apalike}
\bibliography{references}

\end{document}